\documentclass{article}
\usepackage{spconf,amsmath,graphicx}
\usepackage{amsfonts,amssymb}
\usepackage{algorithm}
\usepackage[noend]{algpseudocode}
\usepackage{caption}
\usepackage{listings}
\usepackage[x11names]{xcolor}
\usepackage{color}
\usepackage{xspace}
\usepackage{bbm}
\usepackage{dsfont}
\usepackage[percent]{overpic}
\usepackage{xkeyval}
\usepackage{bm}

\makeatletter
\def\BState{\State\hskip-\ALG@thistlm}
\makeatother

\renewcommand{\Re}[1]{\mbox{$\mathbbm{R}^{#1}$}}

\newdimen\origiwspc
\newdimen\origiwstr
\origiwspc=\fontdimen2=\fontdimen3[figure]{font=small,skip=2pt}

\def\jcloss{\mbox{$\mathcal{L}_{JC}$}}
\def\jloss{\mbox{$\mathcal{L}_{J}$}}
\def\celoss{\mbox{$\mathcal{L}_{CE}$}}

\def\etal{\emph{et al.\xspace}}
\def\ie{\emph{i.e.\xspace}}

\title{$\bm{J}$ Regularization Improves Imbalanced Multiclass Segmentation}
\name{{\parbox[c]{0.9\textwidth}{\centering Fidel A. Guerrero Pe\~{n}a$^{1,5*}$, Pedro D. Marrero Fernandez$^1$, Paul T. Tarr$^{2,3}$, Tsang Ing Ren$^1$, Elliot M. Meyerowitz$^{2,3}$, Alexandre Cunha$^{4,5*}$\\
  \thanks{\footnotesize\fontdimen2\font=0.35ex
We thank financial support from the Brazilian funding agencies
  FACEPE, CAPES and CNPq (FAG, PF, TIR), from the Beckman Institute at Caltech
    to the Center for Advanced Methods in Biological Image Analysis (AC,FAG), 
    from the Howard Hughes Medical Institute (PTT,EMM), and
    thank the IBM Matching Grants Program for computer donation (AC).
    $^*$Corresponding authors: \texttt{fagp@cin.ufpe.br,cunha@caltech.edu.}}}}}

\address{
$^1$Centro de Inform\'atica, Universidade Federal de Pernambuco, Brazil\\
$^2$Howard Hughes Medical Institute, USA\\ $^3$Division of Biology and Biological Engineering,
$^4$Center for Data-Driven Discovery,\\ 
$^5$Center for Advanced Methods in Biological Image Analysis, 
California Institute of Technology, USA
}

\begin{document}
\ninept
\maketitle
\begin{abstract}
  
  We propose a new loss formulation to further advance the multiclass
  segmentation of cluttered cells under weakly supervised conditions.
  We improve the separation of touching and immediate cells, obtaining sharp
  segmentation boundaries with high adequacy, when we add Youden's $J$ statistic 
  regularization term to the cross entropy loss.  This regularization
  intrinsically supports class imbalance thus eliminating the necessity of explicitly
  using weights to balance training.  Simulations demonstrate this capability
  and show how the regularization leads to better results by helping advancing
  the optimization when cross entropy stalls.
  We build upon our previous work on multiclass segmentation by adding yet
  another training class representing gaps between adjacent cells.
  This addition helps the classifier identify narrow gaps as
  background and no longer as touching regions.
  We present results of our methods for 2D and 3D images, from bright field to
  confocal stacks containing different types of cells, and we show that they
  accurately segment individual cells after training with a limited number of
  annotated images, some of which are poorly annotated.

\end{abstract}

\section{Introduction}
\label{sec:intro}

\begin{figure}[t!]
  \centering
  \setlength{\tabcolsep}{1pt}
  \begin{tabular}{cccccc}
    \includegraphics[width=.16\columnwidth]{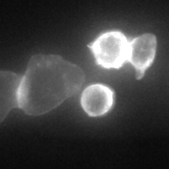}&
    \includegraphics[width=.16\columnwidth]{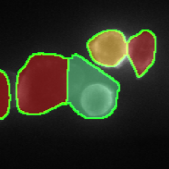}&
    \includegraphics[width=.16\columnwidth]{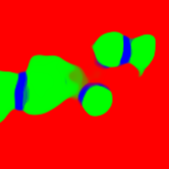}&
    \includegraphics[width=.16\columnwidth]{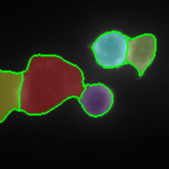}&
    \includegraphics[width=.16\columnwidth]{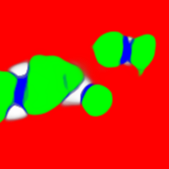}&
    \includegraphics[width=.16\columnwidth]{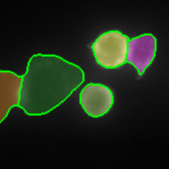}\\[-4.5mm]
    \color{white} \tiny\bf Image & \color{white}\tiny\bf Annotation & \color{white}\tiny\bf $\bm{J3}$ Probability& \color{white}\tiny\bf $\bm{J3}$ Segmentation& \color{white}\tiny\bf $\bm{J4}$ Probability & \color{white}\tiny\bf $\bm{J4}$ Segmentation\\[1mm]
  \end{tabular}
  \caption{\footnotesize A poorly annotated test image is segmented with our
    $J$ regularized loss, \jcloss, using three, $J3$, and four, $J4$, semantic classes. We obtain a
    correct segmentation when training with an added fourth class representing
    gaps and cavities -- predicted white patches shown on $J4$ Probability. The
    result is superior to the annotation, which, unintentionally, missed
    tracing the right contour on the round central cell above. The training of
    our models has been consistently robust despite the presence of weakly
    annotated images, which are present in the total training data.
    Background (red), cell (green), touching (blue), and gap (white) are the
    four classes predicted by our models.
  }
  \vspace{-2mm}
  \color{lightgray}\rule{1.0\columnwidth}{0.2mm}
    \label{fig:mcc3_4}
    \vspace{-8mm}
  \end{figure}

The long-term goal of our work has been the automatic segmentation of cells
found in different modalities of microscope images so that it can ultimately
help in the quantification of biological studies (see e.g.
\cite{guerrero2019weakly,guerrero2018multiclass,cunha2012computational,cunha2010segmenting}).
The task remains a challenge particularly when cells are densely packed in clusters
exhibiting a range of signals and when training with a small number of weak
annotations (see Fig.\ref{fig:mcc3_4}). Separation of cluttered cells is
especially difficult when shared edges have low contrast and are similar to cell
interiors. Weak annotations, when incomplete and inaccurate, can harm the
learning process as the optimizer might be confused when deciding if annotated
and non-annotated regions with same patterns must be segmented or not. Our
proposed solutions aim to resolve these problems with advances in loss
formulation, class imbalance handling, multiclass classification, and data
augmentation.

We propose a new deep learning multiclass segmentation method which classifies
pixels into four distinct classes -- background, cell, touching, and gap -- by
minimizing a loss function that penalizes both cross entropy and Youden's $J$
statistic.  Pixels and voxels classified as touching and gap become either cell or
background in a post-processing step, producing a final segmentation
containing a single mask for each individual cell in the image.

We build upon our recent work 
\cite{guerrero2019weakly, guerrero2018multiclass} to further improve 
multiclass cell segmentation. The introduction of a fourth class, named gap, and of a
new loss lead to better segmentations where small regions separating nearby
cells are now correctly classified as background regions. Slim cell protrusions
are also correctly classified thanks to the balancing offered by our proposed
loss.

  {\bf Previous work}. Recent modeling of new loss functions for segmentation
  \cite{guerrero2019weakly,milletari2016v,salehi2017tversky,berman2018lovasz}
  incorporates a differentiable surrogate over a known performance measurement.
  Unfortunately these are not sufficient to cope with high data imbalance
  typical when segmenting biomedical images. In \cite{kervadec2019} the authors
  review regional losses and propose a contour based loss as an alternative to
  combat imbalance. The work of Brosch \etal\ \cite{brosch2015deep} bears
  similarities to ours as they model their loss as a linear combination of
  sensitivity and specificity measures. But they use mean square errors instead
  and recommend heavily weighting specificity, 95\%, in detriment to
  sensitivity, 5\%, which we believe goes against the importance of equally
  balancing both measures. Sudre \etal\ \cite{sudre2017generalised} proposed
  using the generalized dice overlap introduced in \cite{crum2006generalized}
  as a loss function to avert imbalance in segmentation. Imbalance is achieved
  by explicitly weighting classes as in \cite{ronneberger2015u} but now inversely
  proportional to the square number of pixels. From our experience, this works
  to isolate cell clusters but it is not enough to isolate cells in a
  cluster.

  Pixel weights have been adopted as
  a strategy to balance data \cite{ronneberger2015u,brosch2015deep} including shape aware
  weights \cite{guerrero2018multiclass}. While advantageous they are not
  sufficient to fully separate packed cells or resolve fine details. Equibatches
  \cite{berman2018lovasz} is yet another balancing strategy for segmentation. It
  forces training examples from all classes to be present during every training
  iteration. Multiclass deep learning training for cell segmentation is adopted in
  \cite{guerrero2018multiclass} for 2D images and in \cite{eschweiler2019cnn}
  for 3D confocal stacks.

 \section{Method}
\label{sec:pmethod}
\vspace{-2mm}

{\bf Notation}. 
The goal of panoptic segmentation is to assign to each pixel or voxel
$p\in\Omega$ of a single channel image $x\colon\Omega\subset\Re{d}\to\Re{+}$ a
semantic label, and an instance label when $p$ belongs to a countable
category \cite{kirillov2019panoptic}.
For learning a segmentation we are given a
training set $S=\{(x_i,g_i)\}$ where for every image $x_i$ we know its
ground truth segmentation $g_i$.
In general, we have $g\colon\Omega\to\{0,\ldots,m\}$,
a mapping where $g(p)=0$ for $p$ in the
background and $0 < g(p) \leqslant m$ is a unique label for each object in the
image. Our task is cast as a semantic segmentation problem by modifying
the approach proposed in \cite{guerrero2019weakly} to transform the
instance annotation $g$ into a semantic ground truth $h$, generalizing to
high dimensions by using a $(2k+1)^d$ neighborhood $\eta_k(p),k\geqslant 1$.
Let $y\colon\Omega\to\Re{C+1}$ be the one hot representation for the
$C$-classes in the semantic mapping $h\colon\Omega\to\{0,\ldots,C\}$, and
$n_l=\sum_{p\in\Omega} y_l(p)$ the number of elements of class $l$. We call
$\varrho_{e}\colon\Omega\to\Re{+}$ the bottom hat transform over $g$ using
structuring element $e$, a hyper-sphere whose size is data dependent.
The output of our trained network is a probability map $z$ such that
$z(p)\approx y(p),\forall p$. A post-processing similar to the one
proposed in \cite{guerrero2019weakly} is then applied to build a panoptic
segmentation $\hat{g}$ from $z$.

\textbf{Gap class}.  We have previously shown that using three semantic classes, namely
image background, cell interior, and touching region, increases 
the network discriminative power when segmenting cluttered cells
\cite{guerrero2018multiclass,guerrero2019weakly}. However, misclassified
background regions persisted in some cases, see Fig.\ref{fig:mcc3_4}. 
We speculate this is due to losing background information when merging nearby
cells in the U-Net contracting path, information which is not fully recovered
in the up--sampling path.
By introducing a new training class representing the gap between nearby cells,
the network can now classify the regions separating nearby cells as background.
We name this new class, not surprisingly, gap -- white pixels shown in $J4$,
Fig.\ref{fig:mcc3_4}. These regions are obtained using the bottom hat
transform.  Given an instance annotation $g$, a semantic ground truth $h$ of
our four classes is defined as
\begin{equation*}\label{eq:gtoh}
    h(p) = \left\{ \begin{aligned}
    0 &\;\; \text{if } g(p)=0 \text{ and }\varrho_e(p)=0 - {\scriptsize\text{\em background}}\\
    3 &\;\; \text{if } g(p)=0 \text{ and }\varrho_e(p)>0 - {\scriptsize\text{\em gap}}\\
    2 &\;\; \text{if } g(p^\prime)\neq g(p) \text{ and}\ g(p^\prime)\neq 0,\forall {p^\prime\in\eta_k(p)} - {\scriptsize\text{\em touching}}\\
    1 &\;\; \text{otherwise} - {\scriptsize\text{\em cell}}\\
    \end{aligned} \right.
\end{equation*}
If $p$ is in the background and lies in the bottom hat transform, then $p$ is a gap pixel/voxel, $h(p)=3$. We use $k=2$ in our experiments.

\subsection{$\bm{J}$ regularization}

The {\em J} statistic was formulated by statistician William J. Youden to
improve rating the performance of diagnostic tests of diseases
\cite{youden1950index}. A high index $J$ for a test would imply that this test
could predict with high probability if an individual was diseased or not.  An
ideal test would be able to eliminate false negatives (sick, at risk
individuals falsely reported as healthy) and false positives (healthy
individuals falsely reported as sick) thus always reporting with certainty
diseased (true positive) and healthy (true negative) individuals. Youden
modeled $J$ as the average success of a test on reporting the proportions of
diseased and healthy individuals.  The effectiviness of
this index in binary classification is due to the equal importance it gives to
correctly classifying the subjects belonging and {\em not} belonging to a
class, giving equal weight to true positive (sensitivity) and true negative
(specificity) rates. $J$ is thus a suitable measure for predicting segmentation
with our imbalanced classes: we typically have $n_0\gg n_1\gg n_2 \approx n_3$,
\ie\ touching and gap classes are comprised of a few pixels/voxels when compared to
background and cell classes.
We can write $J = sensitivity + specificity - 1 = TPR + TNR - 1$.
We thus have $J\in [-1,1]$, and we aim to penalize negative
correlations \cite{shan2015improved} and obtain a high $J$ after training.

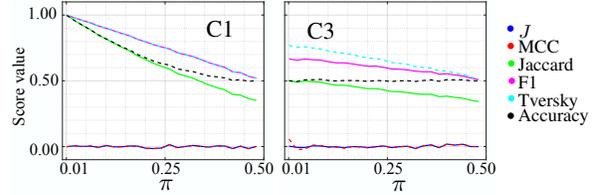
\begin{figure}[t]
    \setlength{\tabcolsep}{1pt}
    \begin{tabular}{cc}
    \begin{overpic}[height=.3\columnwidth]{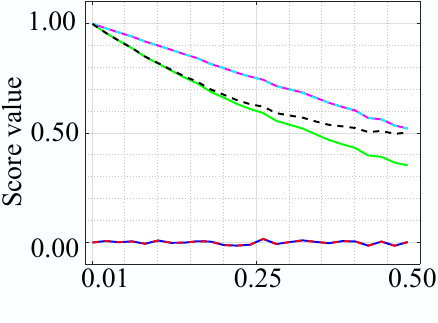}
    \put (54,2) {{ $\pi$}}
    \put (72,60) {{ C1}}
    \end{overpic}    &
    \begin{overpic}[height=.3\columnwidth]{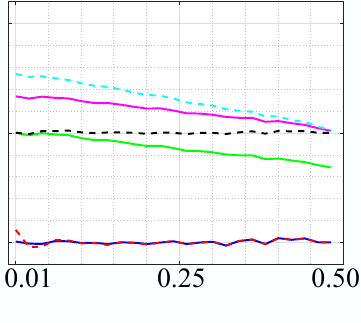}
    \put (10,72) {{ C3}}
        \put (105,35) {{\includegraphics[height=.07\textwidth]{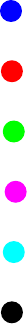}}}
        \put (110,72.5) {{\scriptsize $J$}}
        \put (110,65.5) {{\scriptsize MCC}}
        \put (110,57.5) {{\scriptsize Jaccard}}
        \put (110,49) {{\scriptsize F1}}
        \put (110,41)   {{\scriptsize Tversky}}
        \put (110,34) {{\scriptsize Accuracy}}
        \put (49,2) {{ $\pi$}}
    \end{overpic}  \\
    \end{tabular}
    \caption{\footnotesize Performance of classifiers C1 and C3
      \cite{boughorbel2017optimal} measured by Youden's $J$, Matthews
      Correlation Coefficient (MCC), Jaccard, F1 (Dice), Tversky, and Accuracy scores
      for different imbalance ratios $\pi$. Youden and MCC are the only ones
      almost completely invariant to all imbalance ratios.}
    \vspace{-2mm}
    {\color{lightgray}\rule{1.0\columnwidth}{0.2mm}}
    \vspace{-9mm}
    \label{fig:C1C3}
\end{figure}

We borrow ideas from \cite{boughorbel2017optimal} to compare $J$ to other
popular measures used in loss surrogates \cite{milletari2016v,
  salehi2017tversky, berman2018lovasz}. Note that the most common surrogate for
Accuracy is the Cross Entropy loss \cite{goodfellow2016deep}.  Classifier C1 is
a random prediction where each class has the same imbalance ratio $\pi$ as in
the ground truth. C3 is a random prediction with uniform distribution for all
classes, $\pi=0.5$.
As can be seen in Fig.\ref{fig:C1C3} the performance of $J$ under different
imbalance ratios $\pi\in[0.01,0.50]$ is similar to the Matthews
Correlation Coefficient, MCC \cite{matthews1975}, which is well-known to
perform well under highly imbalanced data \cite{boughorbel2017optimal}.
This is not the case for the Jaccard index , F1 (Dice) score, Tversky index,
and Accuracy, as they all report different values for different imbalance ratio
$\pi$. $J$ should thus be favored when training with imbalanced classes.

\begin{figure}[b!]
    \centering
    \setlength{\tabcolsep}{10pt}
    \begin{tabular}{ccc}
    \begin{overpic}[grid=false,height=.20\columnwidth]{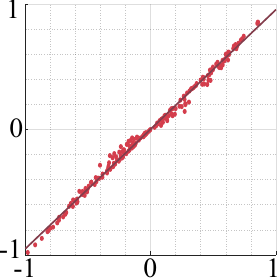}
        \put (40,101) {{\scriptsize $\pi= 0.50$}}
        \put (65,12) {{\scriptsize MCC}}
        \put (11,85) {{\scriptsize $J$}}
    \end{overpic}&
    \begin{overpic}[grid=false,height=.20\columnwidth]{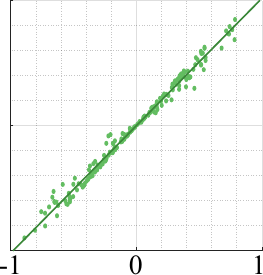}
      \put (40,101) {{\scriptsize $\pi= 0.25$}}
      \put (65,12) {{\scriptsize MCC}}
      \put (9,85) {{\scriptsize $J$}}
    \end{overpic}&
    \begin{overpic}[grid=false,height=.20\columnwidth]{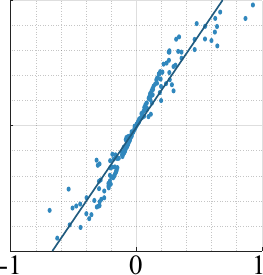}
      \put (40,101) {{\scriptsize $\pi= 0.01$}}
      \put (65,12) {{\scriptsize MCC}}
      \put (9,85) {{\scriptsize $J$}}
    \end{overpic}\\
    \end{tabular}
    \vspace{0mm}
    \caption{\footnotesize Correlation between values of MCC and $J$ for
      different imbalance ratios $\pi$. The linear correlation was measured
      using Pearson Correlation Coefficient, giving values of $0.92\ (\pi=
      0.01), 0.99\ (\pi=0.25), 1.00\ (\pi=0.5)$.}
    \vspace{-2mm}
    \label{fig:MvsMCC}
\end{figure}
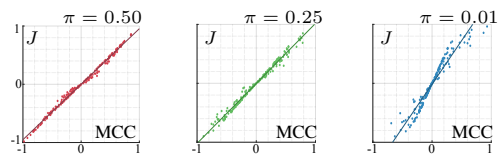

To compare the correlation between $J$ and Matthews Correlation
Coefficient, we used the settings for classifier C3 from
\cite{boughorbel2017optimal}. We then measured the linear correlation between
MCC and $J$ for imbalance ratios $\pi = 0.01, 0.25, 0.50$ by using
Pearson's Correlation Coefficient.  Fig.\ref{fig:MvsMCC} shows an almost
perfect linear correlation for all ratios.  This supports our claim that
Youden's $J$ index is a robust measure for imbalanced binary classification problems.

Assuming a binary segmentation problem, we then define a binary surrogate for $J$ as
\begin{equation}
    \mathcal{L}_J^b(y,z)= -\lambda\log\left(\frac{1+J}{2}\right) = - \lambda \log \left(\frac{\alpha + \beta}{2}\right)
    \label{eq:mcc3}
\end{equation}
with $\alpha$ and $\beta$ soft definitions, respectively,
for TPR and TNR, and $\lambda$ a weighting coefficient.
From Eq. \ref{eq:mcc3}, we define
a multiclass surrogate for $J$ as the sum of pairwise binary surrogates
\begin{equation}
    \mathcal{L}_J(y,z)= - \sum_{i=0}^{C} \sum_{k=0}^{C} \lambda_{i,k} \log \left(\frac{\alpha_{i}+\beta_{i,k}}{2}\right)
    \label{eq:mcc1}
\end{equation}
where $\lambda_{i,k}$ is a pairwise class weight.
$\alpha_i$ and $\beta_{i,k}$
are, respectively, soft definitions for TPR and TNR, where
$i$ is considered to be the positive class and $k$ the negative one. These
definitions are similar to the ones used for Soft Dice \cite{milletari2016v} and
Tversky \cite{salehi2017tversky} loss functions,
\begin{eqnarray*}
\alpha_{i}=  \displaystyle\sum_{p\in\Omega} z_i(p)\cdot \varphi_i(p),& \beta_{i,k}=\displaystyle\sum_{p\in\Omega} (1-z_i(p))\cdot \varphi_k(p)
\end{eqnarray*}
where $\varphi_i(p)=y_i(p)/n_i$.
Inserting these values into Eq.\ref{eq:mcc1} we obtain
\begin{equation}
    \mathcal{L}_J(y,z)= - \sum_{i=0}^{C} \sum_{k=0}^{C} \lambda_{i,k} \log \left(\frac{1}{2}+\sum_{p\in\Omega} z_i(p)\cdot \Delta_{i,k}(p)\right)
    \label{eq:mcc2}
\end{equation}
with $\Delta_{i,k}=(\varphi_i-\varphi_k)/2$.
We use Eq.\ref{eq:mcc2} as a regularizer to cross entropy loss,
$\mathcal{L}_{CE}(y,z)=-\frac{1}{|\Omega|}\sum_{l=0}^{C} \sum_{p\in\Omega}
y_l(p)\cdot \log z_l(p)$, obtaining our training $JC$ loss
$\mathcal{L}_{JC}(y,z)=\mathcal{L}_{CE}(y,z) + \mathcal{L}_J(y,z)$. Of all
solutions with equal values of cross entropy, we favor the one that has the
highest separation between classes. Note that, contrary to
\cite{ronneberger2015u,guerrero2018multiclass}, explicit class weights per pixel
are not used.

\begin{figure}[b!]
    \centering
    \begin{tabular}{c}
    \includegraphics[width=0.8\columnwidth]{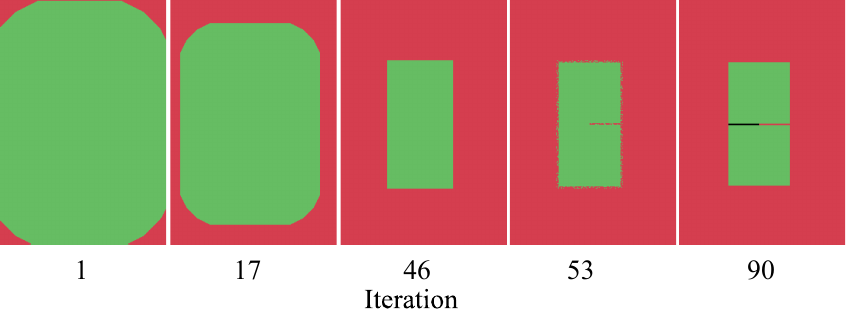}\\
    \end{tabular}
    \setlength{\tabcolsep}{0pt}
    \begin{tabular}{cc}
    \begin{overpic}[width=.5\columnwidth]{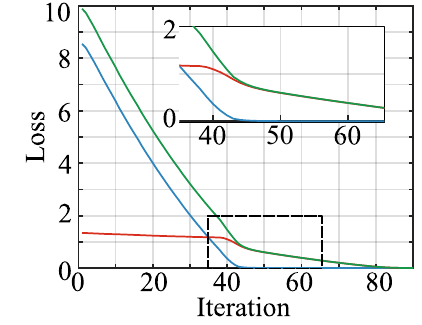}
    \end{overpic}    &
    \begin{overpic}[width=.5\columnwidth]{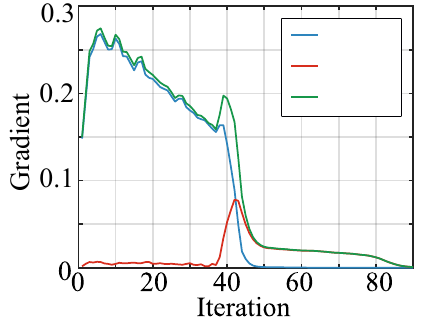}
        \put (76,66.5) {{\scriptsize $\nabla\mathcal{L}_{CE}$}}
        \put (76,59.5) {{\scriptsize $\nabla\mathcal{L}_J$}}
        \put (76,52) {{\scriptsize $\nabla\mathcal{L}_{JC}$}}
    \end{overpic}   \\
    \end{tabular}
    \caption{\footnotesize 
      We simulate segmentation towards ground truth by shrinking an initial
      incorrect segmentation until it shrinkwraps the two target cells (green
      squares above). This happens while we slowly increase the probabilities
      of all pixels towards their correct class. During this stage, \celoss\
      drives optimization.  Around the shrinkwrap point, \jloss\ dominates the
      gradient descent as $\nabla\celoss\approx 0$ is no longer sufficient to
      drive the optimization. The ground truth is achieved (one pixel wide
      notch and touching are identified) thanks to $\nabla\jloss$ which does
      not vanish until the segmentation is correct. {\em Cross entropy and $J$
      statistic work in tandem.  They are not sufficient if used separetely.}}
    \vspace{-2mm}
    \label{fig:cases}
\end{figure}

{\bf Simulation}. We simulate the optimization towards the ground truth to show how the $J$ regularization helps cross entropy, CE, reach the optimum result. The target segmentation consists of two touching square cells separated by a one pixel wide notch covering half of a cell side, see Fig.\ref{fig:cases}. Initially, when the solution is far away ($iter = 1$), CE drives the optimization (large gradients) until it shrinkwraps both cells, at which point ($iter = 46$) its gradient no longer contributes to advance the segmentation. Around that point, $J$ takes over and its gradient is now driving the optimization and it will do so until the optimum is reached. We slowly increase pixel probabilities to its optimal value until we reach ground truth so to mimic real updates. Plots in Fig.\ref{fig:cases} show how the combination of cross entropy and Youden's $J$ statistic work in tandem to achieve the desired result. None would solve the segmentation if considered separately as the vanishing of their gradients would stall the optimization.

\begin{figure}[t!]
    \centering
    \begin{overpic}[width=0.8\columnwidth]{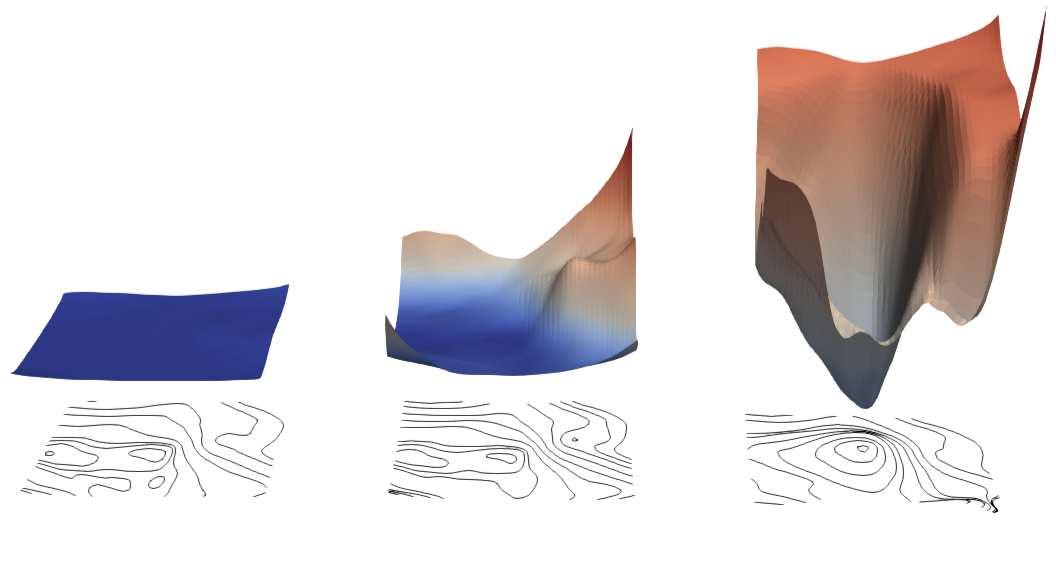}
        \put (12,3) {$\mathcal{L}_{BWM}$}
        \put (47,3) {$\mathcal{L}_{\text{W}^3}$}
        \put (75,3) {$\mathcal{L}_{JC}$}
    \end{overpic}    
    \caption{\footnotesize Loss landscape visualization around a known optimal
      point $\theta^*$ for weighted cross entropy with class balance $\mathcal{L}_{BWM}$
      \cite{ronneberger2015u}, triplex weight map $\mathcal{L}_{\text{W}^3}$
      \cite{guerrero2019weakly}, and our $\mathcal{L}_{JC}$ loss, all in same
      scale. Note how the latter would lead to a faster convergence when close
      to $\theta^*$ due to its steep gradients.}
    \vspace{-2mm}
    {\color{lightgray}\rule{1.0\columnwidth}{0.2mm}}
    \vspace{-9mm}
    \label{fig:losslandscape}
\end{figure}

{\bf Loss visualization}. We use the approach proposed by Li et {\it al.}
\cite{li2018visualizing} to help us visualize how our \jcloss\ loss compares to
others -- $\mathcal{L}_{BWM}$, weighted cross entropy with class balance, and
$\mathcal{L}_{\text{W}^3}$, triplex weight map \cite{guerrero2019weakly} --
around a known optimal point in the optimization space. As shown in
Fig.\ref{fig:losslandscape}, our loss has a cone--like shape whose gradients
favor a fast descent to the optimum, contrary to the other losses
$\mathcal{L}_{BWM}$ and $\mathcal{L}_{\text{W}^3}$ which have near zero
gradients all over potentially preventing the optimization to reach the optimum
-- gradient descent methods are extremely slow to converge in these cases.
Although this analysis is based on a visualization that employs dimensionality
reduction, our evidences from other experiments suggest this behavior spans the
entire optimization space.

\textbf{Gap assignment.}  We obtain a semantic segmentation from the output
probability map $z$ using the Maximum A Posteriori (MAP) decision rule,
$\hat{h}(p)=\arg \max_l z_l(p)$.  A gap pixel $p$, $\hat{h}(p) = 3$, can be
directly classified as a true background pixel or, in case of dubious
probabilities, $z_0(p)\approx z_1(p)\approx z_2(p)$, we assign the second
most likely class to it. This is equivalent of applying MAP on the first three
classes of the output map, $\hat{h}(p)= \arg \max_{l\in \{0,1,2\}}
z_l(p)$. An instance segmentation is achieved then by a sequence of labeling
operations on each region in the semantic segmentation map
\cite{guerrero2019weakly}.
 \vspace{-3mm}
\section{Results}
\vspace{-2mm}
\label{sec:resul}

\begin{figure*}[t]
    \centering
    \setlength{\tabcolsep}{0pt}
    \begin{tabular}{c}
         \includegraphics[width=.99\linewidth]{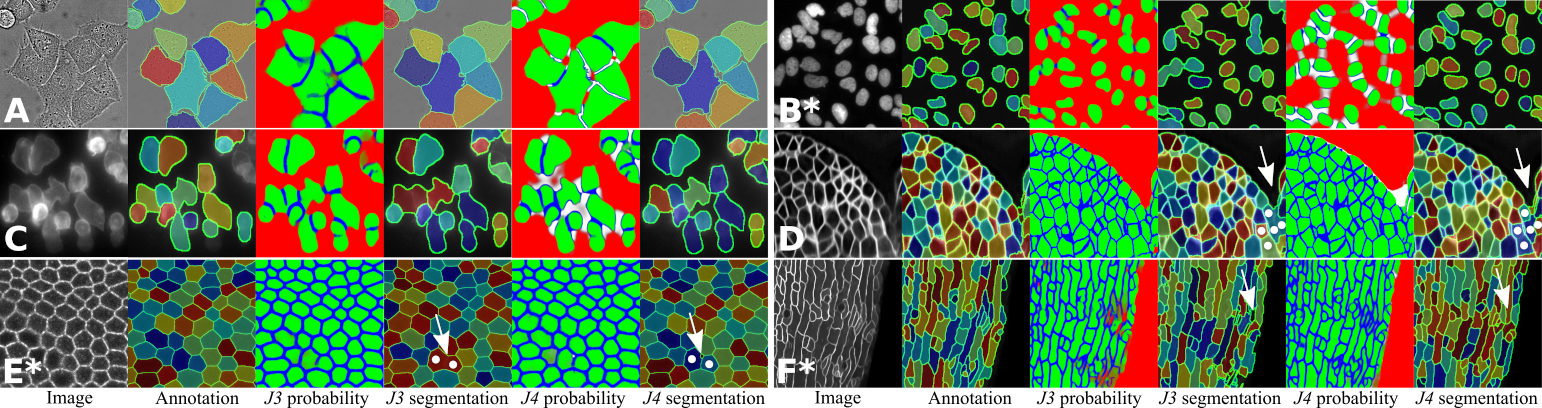}\\
    \end{tabular}
    \caption{\footnotesize Segmentation results for Hela cells (A), Hela {\it
        nuclei} (B), T-Cells (C), {\it Arabidopsis thaliana} meristematic
      cells (a YZ-slice of the 3D segmented stack is shown) (D), Drosophila
      cells (E), and {\it Arabidopsis thaliana} sepal cells (z projection) (F)
      images using networks trained with $J3$ and $J4$ loss functions.
      Probability maps are shown as RGB images with Background (red), Cell
      (green), and Touching (blue) classes. For $J4$, the proximity prediction
      is shown in white. Asterisks (*) indicate zero-shot instance
      segmentations with networks trained exclusively over T-Cells (C).  Colors
      are to show cell separation. Original images were enhanced to help
      visualization. Whites arrows and circles are used to indicate some
      diferences between $J3$ and $J4$.}
    \vspace{-2mm}
    {\color{lightgray}\rule{1.0\textwidth}{0.2mm}}
    \vspace{-7mm}
    \label{fig:others}
\end{figure*}

To facilitate comparing our loss \jcloss\ to losses weighted cross entropy with
class balance (BWM), weighted cross entropy with triplex weight map (W$^3$)
\cite{guerrero2019weakly}, and cross entropy with dice regularization (DSC)
\cite{isensee2019nnu} we use all with the same U-Net
\cite{ronneberger2015u}, with initial weights following a normal distribution 
\cite{glorot2010understanding}, and all equally initialized by fixing all random seeds.
For 3D volumes we used 3D convolutions but maintained the same architecture topology as in 2D
\cite{milletari2016v}.
A Watershed post-processing (WT) is also applied to those results showing weak
touching separation (see \cite{guerrero2019weakly} for details).  The influence
of the gap class over training was also analyzed by comparing $J3$ and $J4$
over a DIC Hela dataset \cite{ISBI2019}, a 3D meristem confocal stack
(see Fig.\ref{fig:meristem}), and T-Cells from \cite{guerrero2019weakly}.  Zero shot
segmentation of Hela cells \cite{ljosa2012annotated} was obtained by using a
model trained over the T-Cells data. We used the optimizer Adam
\cite{kingma2014adam} with initial learning rate of $10^{-4}$. Data
augmentation included random rotation, mirroring, gamma correction, touching
contrast modulation \cite{guerrero2019weakly}, and warping. 
Precision (P05) and F1 score (RQ) were used for cell detection
rates. Segmentation Quality (SQ) and Panoptic Quality were, respectively, used
for measuring contour adequacy and instance segmentation quality
\cite{kirillov2019panoptic}.

\textbf{Instance segmentation performance:} Table \ref{tab:res1} shows a
performance comparison of networks trained with different loss functions. 
Watershed (WT) post-processing effectively increased the
performance of BWM, DSC and W$^3$ when compared with Maximum a Posteriori
(MAP). However, the WT method depends on carefully choosing two parameters.
Networks trained with the proposed \jcloss\
loss are able to improve instance detection rates using only
the parameter-free MAP post-processing. This is due to improvements in the probabilities of gap and touching regions
leading to better cell separation. Because we have a weakly annotated ground truth (see annotation in Fig.
\ref{fig:mcc3_4}), we found SQ values are not always reliable.
\begin{table}[t!]
\begin{tabular}{lccccc}
Loss function   &Post&  P05  & RQ    & SQ    & PQ    \\\hline\hline
BWM             &MAP& 0.6756& 0.5580& 0.8674& 0.4858\\
DSC             &MAP& 0.9028& 0.7674& \textbf{0.9011}& 0.6923\\
W$^3$           &MAP& 0.7384& 0.6305& 0.8721& 0.5513\\\hline
BWM             &WT & 0.8193& 0.8405& 0.8831& 0.7437\\
DSC             &WT & 0.8726& 0.8269& 0.8925& 0.7390\\
W$^3$           &WT & 0.9028& 0.8775& 0.8995& 0.7896\\\hline
$J3$ (Ours)     &MAP& 0.9127& 0.9069& 0.8733& 0.7921\\
$J4$ (Ours)     &MAP& \textbf{0.9334}& \textbf{0.9353}& 0.8689& \textbf{0.8132}\\
\end{tabular}
\caption{\footnotesize Performance comparison of networks trained over Weighted Cross Entropy with class Balance (BWM), Cross Entropy with Dice regularization (DSC) \cite{isensee2019nnu}, Weighted Cross Entropy with Triplex weight map (W$^3$) \cite{guerrero2019weakly}, and \jcloss\ over three, $J3$, and four, $J4$, classes.}
\label{tab:res1}
\vspace{-4mm}
\end{table}

We use \jcloss\ to assess the gap class influence.
Table \ref{tab:res2} shows results obtained over each
dataset. The best Panoptic Quality, PQ, for all cases was obtained with four
classes. An improvement on the Segmentation Quality is observed for the first
two datasets, as a direct consequence of using a fourth class (see first row in
Fig.\ref{fig:others}). However, as stated before, weak annotations in the
case of T-Cells and meristem datasets tainted SQ values: in reality, a visual inspection shows $J4$ offers a better contour adequacy.
The second row of Fig.\ref{fig:others} shows examples of $J3$ and $J4$ 
segmentation and probability maps for T-Cells and meristem volume. Results showed in Figure \ref{fig:others}B, E and F
were obtained with a network trained over T-Cells images (zero-shot instance segmentation).

\begin{table}[t!]
\begin{tabular}{llccc}
Loss function   &Dataset    & RQ    & SQ    & PQ    \\\hline\hline
$J3$            &DIC        & \textbf{0.8950}& 0.8547& 0.7633\\
$J4$            &DIC        & 0.8884& \textbf{0.8833}& \textbf{0.7841}\\\hline
$J3$            &HELA*       & 0.8527& 0.8475& 0.7237\\
$J4$            &HELA*       & \textbf{0.9046}& \textbf{0.8574}& \textbf{0.7764}\\\hline
$J3$            &TCELLS     & 0.9069& \textbf{0.8733}& 0.7921\\
$J4$            &TCELLS     & \textbf{0.9353}& 0.8689& \textbf{0.8132}\\\hline
$J3$            &MERISTEM 3D& 0.8829& \textbf{0.8820}& 0.7787\\
$J4$            &MERISTEM 3D& \textbf{0.8947}& 0.8804& \textbf{0.7878}\\
\end{tabular}
\caption{\footnotesize Results obtained over different datasets show the benefits of using the additional gap class. In all cases a higher PQ value is obtained for $J4$. A (*) indicates zero-shot segmentation.}
\label{tab:res2}
\vspace{-4mm}
\end{table}

\begin{figure}[b!]
\begin{minipage}[t]{1.0\columnwidth}
  \centering
  \includegraphics[width=1.0\columnwidth]{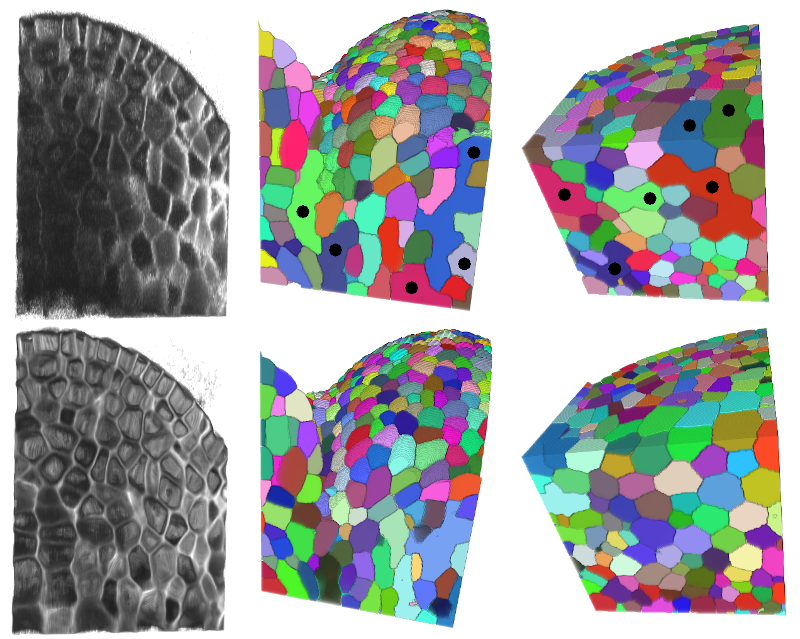}
\end{minipage}
\caption{\label{fig:meristem}\footnotesize {\bf $\mathbf{J4}$ 3D segmentation
    of {\emph{\textbf{Arabidopsis thaliana}}} meristems}. Original and enhanced
  versions (left column) of a portion of a shoot apical meristem image stack
  and their respective segmentations (two views on the middle and right
  columns). Due to space limitation we show only results for this portion which
  has been previously carefully segmented using the watershed with markers
  technique, which we consider as an approximate ground truth.  Enhancing the
  signal quality improves segmentation, as shown for those undersegmented
  regions of the noisy stack manually marked with black circles. Our trained
  network can process large, 1024x1024x508, meristem stacks in under 9 minutes
  using 2 Nvidia K80 GPU cards (31 minutes using a single card).
  Visualizations were prepared using ImageJ 3D Viewer plugin \cite{schmid2010high}.}
  \vspace{-4mm}
    \label{fig:meristem}
\end{figure}

 \vspace{-3mm}
\section{Conclusions}
\label{sec:conclusion}
\vspace{-2mm}

We proposed $JC$ loss, a Youden's $J$ statistic regularization to the bare
cross entropy loss. We build upon our previous work and introduced a new
pixel/voxel class we call {\em gap} which improves classification and contour
adequacy. The approach improved 2D and 3D instance segmentation of highly
cluttered cells even after training with weak annotations. Landscape analysis
and performance evaluation with different loss functions suggest our new loss
is superior to segment cluttered cells.  In future work we plan to optimize
the proposed pairwise loss to be linear in the number of classes and
extensively compare our methods using benchmarks.
 
\bibliographystyle{IEEEbib}

\end{document}